\documentclass{article}
\usepackage{adjustbox}
\usepackage{multirow}
\usepackage{booktabs}
\usepackage{microtype}
\usepackage{graphicx}
\usepackage{subfigure}
\usepackage{booktabs} 

\usepackage{hyperref}

\usepackage[accepted]{icml2024}

\usepackage{amsmath}
\usepackage{amssymb}
\usepackage{mathtools}
\usepackage{amsthm}
\usepackage{hyperref}

\usepackage[capitalize,noabbrev]{cleveref}

\theoremstyle{plain}

\theoremstyle{definition}

\theoremstyle{remark}

\usepackage[textsize=tiny]{todonotes}

\icmltitlerunning{DistilDIRE: A Small, Fast, Cheap and Lightweight Diffusion Synthesized Deepfake Detection}

\begin{document}

\twocolumn[
\icmltitle{DistilDIRE: A Small, Fast, Cheap and Lightweight Diffusion Synthesized Deepfake Detection}



\icmlsetsymbol{equal}{*}

\begin{icmlauthorlist}
\icmlauthor{Yewon Lim}{equal,yonsei,purdue,miraflow}
\icmlauthor{Changyeon Lee}{equal,yonsei,purdue,miraflow}
\icmlauthor{Aerin Kim}{miraflow,truemedia}
\icmlauthor{Oren Etzioni}{truemedia}
\end{icmlauthorlist}

\icmlaffiliation{yonsei}{Department of Computer Science and Engineering, Yonsei University, Seoul, Republic of Korea}
\icmlaffiliation{purdue}{Computer and Information Technology, Purdue University, West Lafayette IN, United States}
\icmlaffiliation{miraflow}{Miraflow, Kirkland WA, United States}
\icmlaffiliation{truemedia}{TrueMedia.org, Seattle WA, United States}

\icmlcorrespondingauthor{Aerin Kim}{aerin@miraflow.ai}

\icmlkeywords{Diffusion, Deepfake, Foundation Models, Efficient Deep Learning}

\vskip 0.3in
]



\printAffiliationsAndNotice{\icmlEqualContribution} 

\begin{abstract}
A dramatic influx of diffusion-generated images has marked recent years, posing unique challenges to current detection technologies. While the task of identifying these images falls under binary classification, a seemingly straightforward category, the computational load is significant when employing the “reconstruction then compare” technique. This approach, known as DIRE (Diffusion Reconstruction Error), not only identifies diffusion-generated images but also detects those produced by GANs, highlighting the technique's broad applicability. To address the computational challenges and improve efficiency, we propose distilling the knowledge embedded in diffusion models to develop rapid deepfake detection models. Our approach, aimed at creating a small, fast, cheap, and lightweight diffusion synthesized deepfake detector, maintains robust performance while significantly reducing operational demands. Maintaining performance, our experimental results indicate an inference speed 3.2 times faster than the existing DIRE framework. This advance not only enhances the practicality of deploying these systems in real-world settings but also paves the way for future research endeavors that seek to leverage diffusion model knowledge. The code and weights for our framework are available at \url{https://github.com/miraflow/DistilDIRE}.

\end{abstract}
\section{Introduction}
From generating high-fidelity images \cite{rombach2022high,karras2017progressive,stylegan, saharia2022photorealistic} to synthesizing complex videos \cite{videoworldsimulators2024, bartal2024lumiere, gao2024vista} and musical compositions \cite{Huang2023Noise2MusicTM}, and deepfakes \cite{chen2023textimage, bhattacharyya2024diffusion}, diffusion models \cite{ho2020denoising, song2022denoising} have demonstrated exceptional capabilities in various domains by leveraging stochastic diffusion processes.


Current deepfake detectors struggle to distinguish unseen diffusion-generated images.  To address this, researchers have developed a new method that uses a pre-trained diffusion model to more accurately reconstruct these images. This method uses the error between an input image and its reconstructed version, known as the DIRE \cite{wang2023dire} representation, as input for a binary classifier to detect deepfakes. This `reconstruction then compare' approach has achieved state-of-the-art results on the DiffusionForensic dataset \cite{wang2023dire}, which includes both GAN~\cite{goodfellow2020generative} generated and diffusion-generated images.

The DIRE framework, although it works well, is too slow for practical use because it needs to compute the diffusion trajectory during training and inference. We address this limitation by proposing a distilling approach that uses a fraction of the information from pretrained diffusion model and predefined teacher classifier model. Specifically, We leverage the DIRE feature extracted by a ImageNet pre-trained ResNet-50 classifier (\emph{teacher}) to train a binary classifier (\emph{student}) for deepfake detection. Additionally, we incorporate the first-time step noise from a pre-trained diffusion model with the original image, effectively conveying the image distribution as interpreted by the diffusion.

This method has led to the development of `\emph{\textbf{DistilDIRE}}', a small, fast, cheap and lightweight deepfake detection model suitable for real-world applications.
\section{Related Work}
\label{sec:related}
Our work aims to develop a light and fast diffusion synthesized deepfake detection model with only fraction of diffusion trajectory information. 

\subsection{Denoising Diffusion Implicit Models}
Diffusion models introduce a novel approach to generative modeling by leveraging inverse transformations to progressively reduce noise in data generation. While DDPM~\cite{ho2020denoising} constructs the generation process using Marcov chains and thus require sequential sampling without skipping, DDIM~\cite{song2022denoising} can generate images faster than DDPM due to its unique design without Marcov Chain. It models reverse process as follows.
\begin{equation}
\begin{aligned}
    \text{x}_{t-1} = \sqrt{\alpha_{t-1}} \left(\frac{\text{x}_t - \sqrt{1 - \alpha_t} \epsilon_\theta(\text{x}_t, t)}{\sqrt{\alpha_t}}\right) \\
    + \sqrt{1 - \alpha_{t-1} - \sigma_t^2} \cdot \mathbf{\epsilon}_\theta(\text{x}_t, t) +\sigma_t \epsilon_t\label{eq:ddim_reverse}
\end{aligned}
\end{equation}

When $\sigma_t $ goes 0, \cref{eq:ddim_reverse} becomes deterministic from which we can approximately revert image to noise, which is called DDIM inversion. Specifically, DDIM inversion is performed as follows.
\begin{equation}
\begin{aligned}
    \frac{\text{x}_{t+1}}{\sqrt{\alpha_{t+1}}}  = \frac{\text{x}_t}{\sqrt{\alpha_t}}  + \left(\sqrt{\frac{1 - \alpha_{t+1}}{\alpha_{t+1}}} - \sqrt{\frac{1 - \alpha_{t}}{\alpha_t}}\right) \epsilon_\theta(\text{x}_t, t). \label{eq:ddim_forward}
\end{aligned}
\end{equation}

This process aims to obtain the corresponding noisy sample $\text{x}_T$ for an input image $\text{x}_0$. However, inverting or sampling step-by-step is highly time-consuming. To accelerate the diffusion model sampling, DDIM allows sampling a subset of $S$ steps $\tau_1, \dots, \tau_S$, where the neighboring $\text{x}_t$ and $\text{x}_{t+1}$ become $\text{x}_{\tau_t}$ and $\text{x}_{\tau_{t+1}}$, respectively, in \cref{eq:ddim_reverse} and \cref{eq:ddim_forward}.

\subsection{Diffusion Synthesized Image Detection}
As diffusion models advance and generate seamless images, it becomes increasingly challenging to detect these artificially created ones. To tackle this, DIRE~\cite{wang2023dire} offers a solution by leveraging pre-trained diffusion models and extracting their implicit understanding of image distributions. Their research reveals that genuine images exhibit significant reconstruction errors when subjected to DDIM inversion~\cite{song2022denoising}, whereas counterfeit images generated by diffusion models tend to yield lower reconstruction errors.

However, a notable bottleneck arises in terms of inference time, as implementing the DIRE framework requires invoking the ADM model at least 40 times to obtain the DIRE image. 

DNF, or Diffusion Noise Feature~\cite{zhang2024diffusion}, represents another advancement in diffusion-generated image detection. This method delves into the intricate process of diffusion, revealing that the predicted noises generated during the inverse diffusion process can serve as valuable indicators of the underlying image distribution. By exploiting these noise features, deciding whether given image is genuine or counterfeit can be done with higher accuracy.

However, DNF also requires multiple iterations of diffusion model calls before the deepfake probability can be calculated. This highlights the computational complexity involved in such image detection methods and underscores the ongoing efforts to streamline and optimize these processes for practical implementation.
\section{Methodology}
\label{sec:method}
\subsection{Architecture}
\begin{figure}[t]
    \vspace{+6pt}
    \includegraphics[width=\columnwidth]{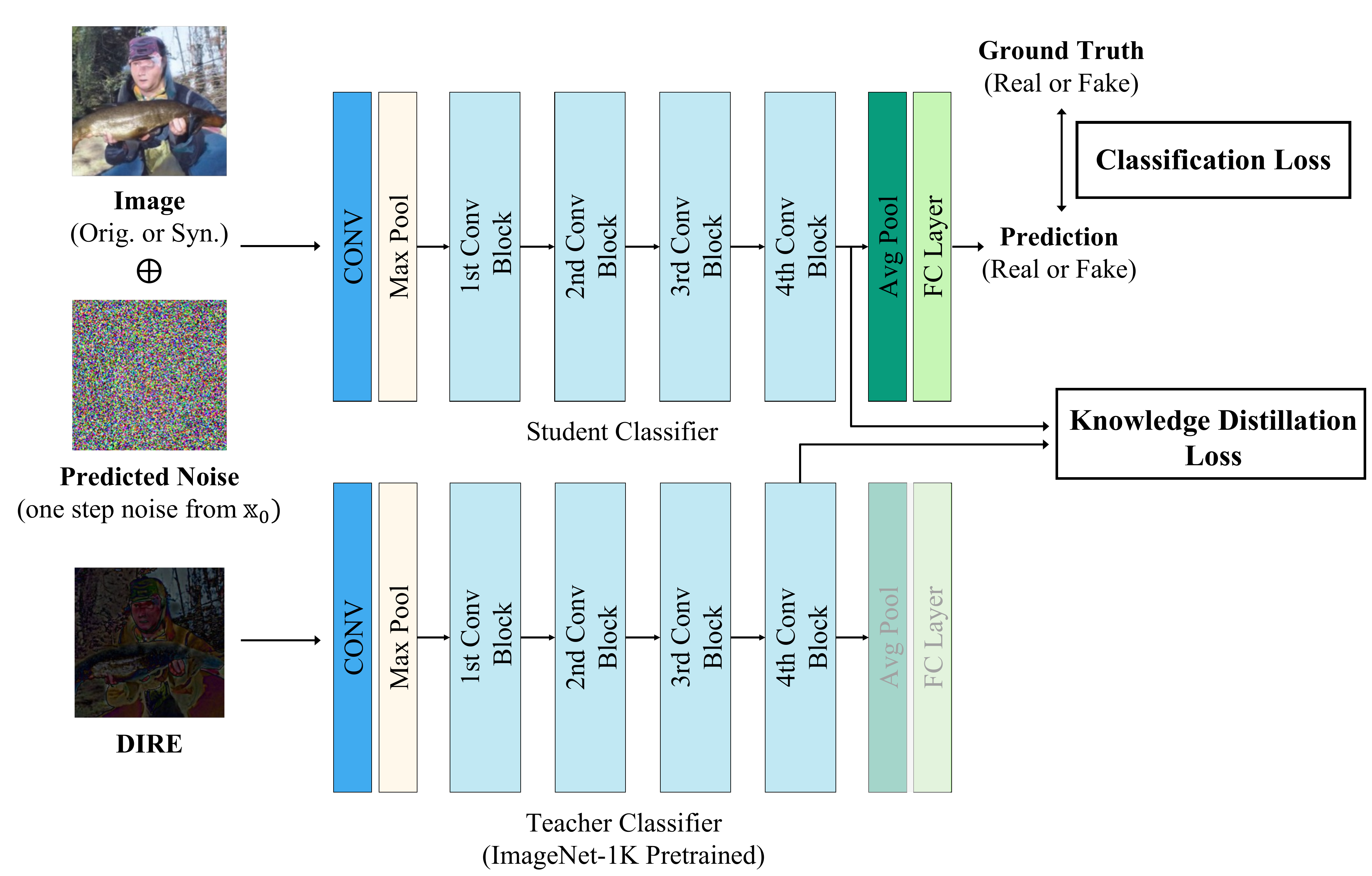}
    \caption{
        Overview of our framework.
        }
    \label{fig:sec3-overview}
\end{figure}

The DistilDire framework, shown in \cref{fig:sec3-overview}, uses a pre-trained ResNet-50 \cite{He_2016_CVPR}, which was initially trained on the ImageNet-1K \cite{deng2009imagenet} dataset, as a teacher model. 

The teacher model, loaded with pretrained weights, remains frozen during training. This model provides a stable reference for the student model, ensuring that the extracted features are robust and generalized. Unlike the teacher model, the student model is trained from scratch. During the training process, the student model simultaneously incorporates classification loss and knowledge distillation loss.

What makes our method unique is that we combine the original image and the predicted noise from that original image into the learning (student) model. Specifically, we utilize noise from the first time step obtained from a pretrained Ablated Diffusion Model (ADM) \cite{dhariwal2021diffusion} as an additional input. The final input to the student model consists of the original image $\textbf{x}$ itself and the noise $\epsilon_{0}$ from the first denoising step. The input to the model is as follows, where $i$ represents the index of the mini-batch. \begin{equation}
    \textbf{x}'_i = \mathsf{cat}(\textbf{x}_i, \epsilon_{0,i})
\end{equation}The student model then predicts whether the input image is real or synthetic (fake).

Knowledge distillation from the teacher model is achieved by extracting feature maps before the classifier head and computing the Mean Squared Error (MSE) loss between these feature maps and those of the student model. The knowledge distillation loss serves as regularization, encouraging the student model's feature maps to mirror the teacher model's without the need for time-intensive inversion and reconstruction.

The student model uses Binary Cross-Entropy Loss (BCE). This helps the model learn how to tell the difference between real and synthetic images. The final way we calculate loss is shown in \cref{eq:loss_equation}, \cref{eq:binary_loss}, and \cref{eq:distil_loss}.

\vspace{-2pt}
\begin{equation}
\begin{aligned}
    \mathcal{L}_\text{total} = \mathcal{L}_\text{classification} +  \lambda \cdot \mathcal{L}_\text{distillation}
    \label{eq:loss_equation}
\end{aligned}
\end{equation}
\vspace{-5.3pt}
\begin{equation}
    \mathcal{L}_\text{classification} = -\frac{1}{N}\sum_{i=1}^N (\mathbf{y}_{i}\mathrm{log}(\mathbf{y}'_{i})+(1-\mathbf{y}_{i})\mathrm{log}(1-\mathbf{y}'_{i}))\label{eq:binary_loss}
\end{equation}

\vspace{-10.8pt}
\begin{equation}
    \mathcal{L}_\text{distillation} = \frac{1}{N}\sum_{i=1}^N \lVert \mathcal{F'}(\mathsf{DIRE}(\textbf{x}_i)) - \mathcal{F}(\textbf{x}'_i) \rVert _2\label{eq:distil_loss}
\vspace{-1mm}
\end{equation}

In the above equations, $\mathbf{y}_i, \mathbf{y}_i'$, $\mathcal{F'}$, and $\mathcal{F}$ represents ground-truth label, model prediction, ImageNet-1K pretrained ResNet-50 feature extractor, and naive ResNet-50 feature extractor, respectively. Here, $N$ is a mini-batch size and $\lambda$ decides how much the distillation process will affect the outcome.

In essence, DistilDIRE uses a pretrained teacher model for generalized feature extraction, while the student model learns its ability to distinguish real from synthetic images through classification and knowledge distillation losses. This approach balances the use of pretrained knowledge and effective learning of new representations.

\subsection{Noise Extraction}
In DistilDIRE, we extract the initial noise \(\mathbf{\epsilon}_0\) to determine whether a given image is generated by a diffusion model. By analyzing the characteristics of \(\mathbf{\epsilon}_0\), we aim to identify signatures that are indicative of diffusion-generated images.

To extract \(\mathbf{\epsilon}_0\), we leverage the DDIM formulation (\cref{eq:ddim_forward}).
By focusing on the case when \(t = 1\), we can isolate the initial noise \(\mathbf{\epsilon}_0\) from the initial image \(\textbf{x}_0\):

\begin{equation}
\begin{aligned}
    \mathbf{\epsilon}_0 = \frac{\textbf{x}_1 - \sqrt{\alpha_1} \hat{\textbf{x}}_0}{\sqrt{1 - \alpha_1}}.
\end{aligned}
\end{equation}

Extracted $\mathbf{\epsilon}_0$ is concatenated with the input image $\textbf{x}_0$ and then forwared to classifier to determine whether given image is synthesized or not.
\begin{table*}[h]
    \Large
    \caption{
    Quantitative evaluation (Accuracy (\%), AP (\%)) of our framework alongside other state-of-the-art deepfake detection methodologies on the ImageNet and CelebA-HQ datasets. `*' denotes the generators present in the training dataset of each subset dataset. ImageNet test subset image generators are ADM~\cite{dhariwal2021diffusion}, Stable Diffusion v1~\cite{rombach2022high} and CelebA-HQ test subset image generators are IF~\cite{IF}, DALL-E 2~\cite{ramesh2022hierarchical}, Stable Diffusion v2~\cite{rombach2022high}, Midjourney~\cite{midjourney2022} \\
    }
    \begin{adjustbox}{width=\textwidth}
        \begin{tabular}{@{}l|c|cc|cccc|c@{}}
    \toprule
    \multirow{3}{*}{Method} & \multirow{3}{*}{Training Dataset}& \multicolumn{2}{c}{ImageNet} & \multicolumn{4}{c}{CelebA-HQ} &  \multirow{2}{*}{Performance}  \\ 
    &  & \multicolumn{2}{c}{Testing Generators} & \multicolumn{4}{c}{Testing Generators} & \multirow{2}{*}{Average}
    \\
    &   & ADM*      & SD-v1     & IF        & DALL-E 2 & SD-v2*   & Midjourney     & 
                                         \\
    \midrule
    \multirow{2}{*}{CNNDet \cite{wang2019cnngenerated}}                  
                                        & DiffusionForensics~-~ImageNet  & 71.6/79.8 & 51.0/51.2 & 36.5/41.2 & 54.2/52.2 & 37.0/41.6 & 48.4/49.1 & 49.8/52.5      \\
                                        & DiffusionForensics~-~CelebA-HQ  & 51.0/58.8 & 52.6/68.0 & 53.6/53.9 & 54.2/52.2 & 78.4/69.9 & 73.6/67.7 & 60.6/61.8      \\    
    \midrule
    \multirow{2}{*}{DIRE \cite{wang2023dire}}               
                                        & DiffusionForensics~-~ImageNet  & 99.8/99.9 & 98.2/99.9 & 50.0/50.0 & 50.0/50.0 & 50.0/50.0 & 50.0/50.0 & 66.3/66.6      \\
                                        & DiffusionForensics~-~CelebA-HQ  & 99.8/99.9 & 58.2/66.2 & 96.8/100 & 93.4/100 & 96.7/100  & 95.0/100 & 90.0/94.4   \\
    
    \midrule
    \multirow{2}{*}{\textbf{Ours (DistilDIRE)}}                  
                                       & DiffusionForensics~-~ImageNet & 98.4/99.5 & 99.0/99.7 & 86.5/95.2 & 89.3/99.2 & 77.3/86.7 & 85.5/88.6 & 89.3/94.8 \\ 
                                        & DiffusionForensics~-~CelebA-HQ  & 61.5/79.4 & 50.2/89.4 & 97.4/99.7 & 93.3/99.1 & 100/100 & 99.6/100 & 83.7/94.6 \\ 
    \bottomrule
\end{tabular}
    \end{adjustbox}
    \label{tab:main_result} 
\end{table*}

\section{Experiment}
\label{sec:exp}
\subsection{Experiments Setup}
\subsubsection{Datasets and Pre-processing}

For the experiments in this study, we utilized the DiffusionForensics dataset \cite{wang2023dire}, incorporating specific subsets from ImageNet and CelebA-HQ \cite{karras2017progressive}.
LSUN-Bedroom~\cite{yu2015lsun} subset is excluded in the experiment for practical reason.

In the training phase, both original images and noise features underwent resizing to 224 x 224, horizontal flipping with a 0.5 probability and normalization. The first time step noises derived from ADM were resized to 224 x 224 as well and horizontally flipped with an equivalent probability to the images and DIRE images.  However, they were not normalized. During testing, images were centered and cropped to a size of 224 x 224.
Different lambda $\lambda$ values for each sub-dataset are tested during the training phase. These values are tuned to optimize the balance between classification loss and knowledge distillation loss, therefore enhancing the model's effectiveness across different types of data. For reporting, $\lambda$ of 0.5 is applied for both ImageNet and CelebA-HQ subset.


\subsubsection{Evaluation}
To evaluate the performance of classifier in categorizing synthetic (fake) deepfake images, we employed metrics including, accuracy, and average precision (AP). The accuracy threshold for computation was set at 0.5. Evaluation was conducted by constructing datasets for each subset of ImageNet and CelebA-HQ, comprising both real images and their corresponding synthetic counterparts.

\subsection{Results}
\subsubsection{Deepfake Detection Performance}
Our evaluation of the DistilDIRE framework demonstrates a meaningful significant advancement in deepfake detection through a balance of high performance and reduced computational demand. The results, as detailed in Table 1, showcase the efficacy of DistilDIRE across varied test conditions on the ImageNet and CelebA-HQ datasets.

In the ImageNet subset, DistilDIRE achieved accuracy and average precision (AP) scores close to the best-performing models, with metrics like 99.0/99.7\% against SD-v1 and 98.4/99.5\% against ADM. For the CelebA-HQ subset, it performed exceptionally well against modern generators such as SD-v2 and Midjourney. These results are particularly noteworthy as they come close to those of the DIRE model, which, while slightly more accurate, requires substantially higher computational resources.

In addition to diffusion model-generated images, our results indicate that DistilDIRE is also highly effective in detecting GAN-generated images. This capability is crucial as GANs are a common method for creating synthetic media. The robust performance across both GAN-generated and diffusion model-generated images signifies the versatility and effectiveness of the DistilDIRE framework.

Compared to traditional models such as CNNDet and even the state-of-the-art DIRE, DistilDIRE not only stands out in terms of detection accuracy but also in computational efficiency. While DIRE provides top-tier detection capabilities, its practical application is hindered by its high computational load. DistilDIRE, on the other hand, offers a compelling alternative by reducing the computational demand by approximately 97\% compared to DIRE, as noted in our inference time and computational efficiency evaluations.

\subsubsection{Inference Time Comparison}

The inference time and computational efficiency of the models were evaluated, specifically focusing on the performance of these models on a single NVIDIA A10 GPU, widely used for deep learning inference in 2024 at the time of writing. The detailed results of this evaluation are presented in \cref{tab:infer_time_results}. The FLOPS was measured using PyTorch FLOPS count implementation \cite{pytorchTorchtntutilsflopsFlopTensorDispatchModex2014}.

The newly proposed DistilDIRE model significantly optimizes computational resources compared to its predecessor, DIRE. While DIRE requires a substantial 149.62 TFLOPS (Tera Floating Point Operations per Second) to process an image, DistilDIRE reduces this demand to just 5.01 TFLOPS, achieving a remarkable reduction by approximately 97\%. Additionally, the inference time for DIRE is around 6.978 seconds per image, which is reduced by approximately 69\% to 2.183 seconds in DistilDIRE (3.2 times faster than DIRE).

CNNDet, another model listed, shows the least computational demand with 0.021 TFLOPS and the shortest inference time at 1.687 seconds per image. This efficiency can be attributed to CNNDet not utilizing the computationally intensive diffusion process used in DIRE and DNF models.

Comparing the DistilDIRE model with DNF, which also employs a diffusion process, DistilDIRE demonstrates improved performance. DNF has a relatively high computational requirement of 20.88 TFLOPS and an inference time of 3.226 seconds per image. In contrast, DistilDIRE, while having a slightly higher computational load than CNNDet, manages to perform faster than DNF, indicating an efficient balance between computational complexity and processing time.

\begin{table}[t]
    \Large
    \vspace{-8pt}
    \caption{Inference Time Comparison for 256 x 256 Images}
    \centering
    \begin{adjustbox}{width=\columnwidth}
    \begin{tabular}{@{}lccc@{}}
    \\
    \toprule
    \text{Method} & \text{Avg. Inference Time (sec/it)} & \# \text{ of Params (M)} & \text{FLOPS (T)} \\ 
    \midrule
    \text{CNNDet \cite{wang2019cnngenerated}} & 1.687 & 23.51 & 0.021 \\ 
    \midrule
    \text{DIRE \cite{wang2023dire}} & 6.978 & 576.32 & 149.62 \\ 
    \midrule
    \text{DNF \cite{zhang2024diffusion}} & 3.226 & 137.18 & 20.88 \\ 
    \midrule
    \textbf{Ours (DistilDIRE)} & 2.183 & 576.33 & 5.01 \\
    \bottomrule
    \end{tabular}
    \end{adjustbox}
    \label{tab:infer_time_results} 

\end{table}

\begin{table*}[h]
    \vspace{-5pt}
    \Large
    \caption{Ablation study on DistilDIRE knowledge distillation (KD) loss on the ImageNet and CelebA-HQ datasets (Accuracy (\%), AP (\%)).\\}
    \begin{adjustbox}{width=\textwidth}
        \begin{tabular}{@{}c|c|cc|cccc|c@{}}
    \toprule
    \multirow{3}{*}{With KD Loss} & \multirow{3}{*}{Training Dataset}& \multicolumn{2}{c}{ImageNet} & \multicolumn{4}{c}{CelebA-HQ} &  \multirow{2}{*}{Performance}  \\ 
    &  & \multicolumn{2}{c}{Testing Generators} & \multicolumn{4}{c}{Testing Generators} & \multirow{2}{*}{Average}
    \\
    & \text{}   & ADM*      & SD-v1     & IF        & DALL-E 2 & SD-v2*   & Midjourney     & 
                                         \\
    \midrule
    \multirow{2}{*}{\checkmark}           
                                       & DiffusionForensics~-~ImageNet & \textbf{98.4/99.5} & \textbf{99.0/99.7} & \textbf{86.5/95.2} & \textbf{89.3/99.2} & \textbf{77.3/86.7} & \textbf{85.5/88.6} & \textbf{89.3/94.8} \\ 
                                        & DiffusionForensics~-~CelebA-HQ  & \textbf{61.5/79.4} & \textbf{50.2/89.4} & \textbf{97.4/99.7} & \textbf{93.3/99.1} & \textbf{100/100} & \textbf{99.6/100} & \textbf{83.7/94.6}\\
    \midrule
    \multirow{2}{*}{}                 
                                       & DiffusionForensics~-~ImageNet & 95.9/99.5 & 96.6/99.5 & 64.7/92.8 & 55.0/98.5 & 55.1/73.9 & 37.7/17.6 & 67.5/80.3 \\ 
                                        & DiffusionForensics~-~CelebA-HQ  & 56.1/67.0 & 48.8/87.0 & 53.2/87.9 & 69.9/67.3 & 99.9/100 & 91.1/36.7 & 69.8/74.3 \\ 
    
    \bottomrule
\end{tabular}
    \end{adjustbox}
    \label{tab:ablation_distil_loss} 
\end{table*}

\subsubsection{Ablation Study}
In this ablation study, we aimed to examine the effect of incorporating knowledge distillation (KD) loss alongside classification loss during the training process. The primary objective was to determine whether the inclusion of KD loss contributes significantly to model performance across two datasets: ImageNet and CelebA-HQ. As shown in \cref{tab:ablation_distil_loss}, the inclusion of KD loss markedly improved Accuracy and Average Precision (AP) for all tested generators. 
Excluding KD loss, by training with only classifcation loss, led to noticeable declines in both Accuracy and AP, particularly with new generators, underscoring the importance of KD loss for optimal performance.

The cross validation analysis indicates that incorporating KD loss not only boosts the Accuracy and AP but also enhances the model's ability to generalize across different generators and datasets. The performance average distinctly favored configurations with KD loss, with a notable 17.9\% and 17.4\% increase in Accuracy and AP respectively when KD loss was used.
\section{Conclusion}
\label{sec:conclusion}

The strength of DistilDIRE lies in its strategic use of knowledge distillation, where the model learns to mimic the decision-making process of a more complex system (DIRE) without engaging in computationally intensive tasks. This approach not only preserves high detection capabilities but also ensures the model is practical for real-time applications. 
DistilDIRE achieves approximately 3.2 times faster speed than existing framework while maintaining high performance.
This combination of fast inference speed and high performance can efficiently handle large inputs such as deepfake videos.



\bibliography{main.bbl}
\bibliographystyle{icml2024}




\end{document}